\documentclass[runningheads]{llncs}

\usepackage{eccv}

\usepackage{eccvabbrv}

\usepackage{graphicx}
\usepackage{booktabs}
\usepackage{multirow}
\usepackage{wrapfig}

\usepackage[accsupp]{axessibility}

\usepackage{hyperref}

\usepackage{orcidlink}

\title{Walk through Paintings: Egocentric World Models from Internet Priors}
\titlerunning{Walk through Paintings: EgoWM from Internet Priors}

\author{Anurag Bagchi$^{1}$ \quad
Zhipeng Bao$^{1}$ \quad
Homanga Bharadhwaj$^{1}$ \quad \\
Yu-Xiong Wang$^{2}$ \quad
Pavel Tokmakov$^{3}$\textsuperscript{*} \quad
Martial Hebert$^{1}$\textsuperscript{*} \\
$^{1}$CMU \quad
$^{2}$UIUC\quad
$^{3}$TRI \\[0.5em]
\href{https://egowm.github.io}{egowm.github.io}}
\authorrunning{A. Bagchi et al.}
\institute{}

\begin{document}

\maketitle
\footnotetext[1]{Equal advising.}

\begin{figure}
  \centering
  \includegraphics[width=\textwidth]{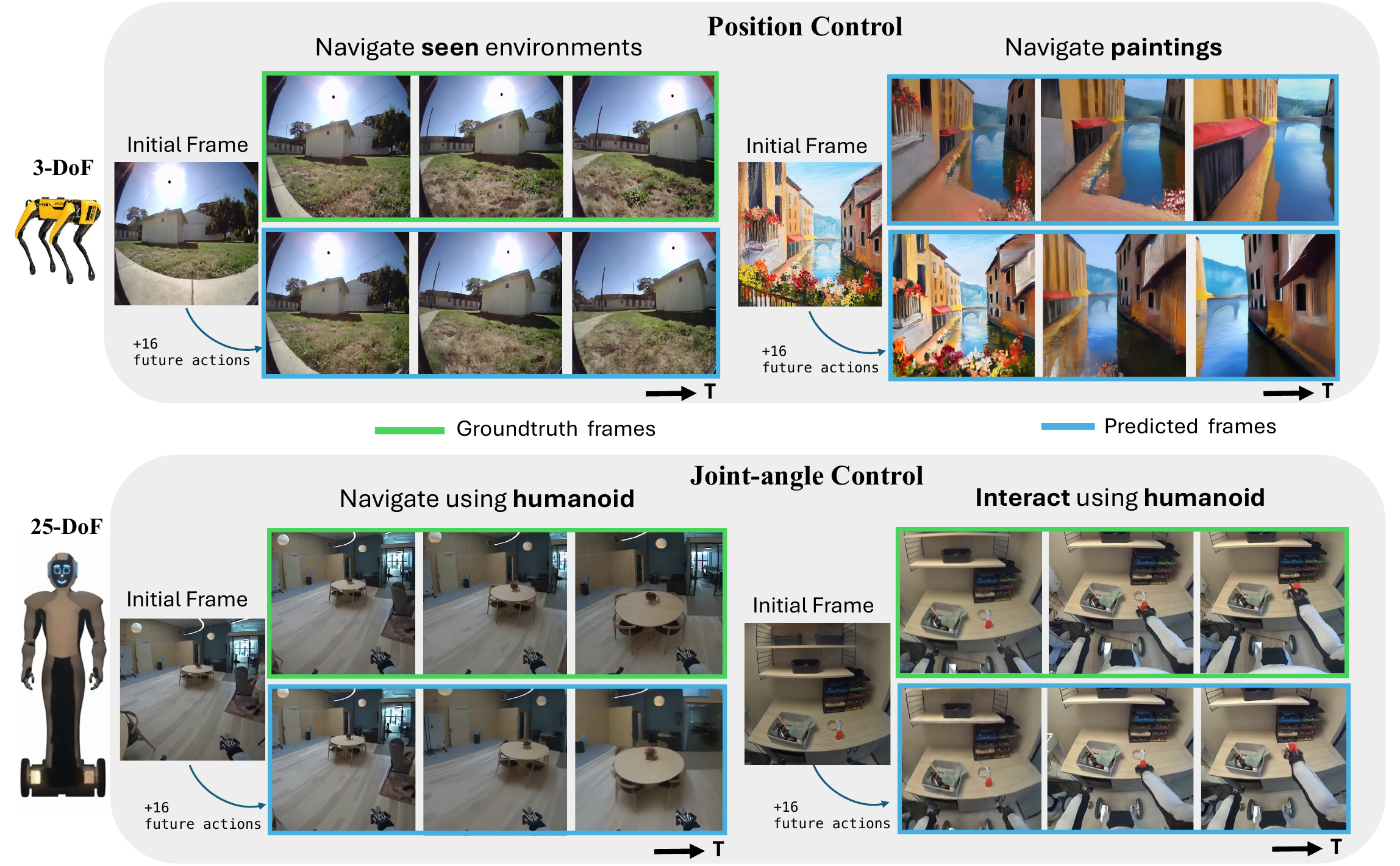}
  \caption{Our framework generates future frame predictions (shown in {\color{blue}blue}) that accurately follow the provided robot actions (ground-truth frames shown in {\color{green}green}). Notably, it effortlessly generalizes to vastly different embodiments, tasks, and domains, including non-realistic ones, such as navigating within paintings (top right). This generalization is enabled by our simple, universal architecture.}
  \vspace{-10px}
  \label{fig:teaser}
\end{figure}

\begin{abstract}
    What if a video generation model could not only imagine a plausible future, but the correct one -- accurately reflecting how the world changes with each action?
We answer this by presenting the Egocentric World Model (EgoWM), a simple, architecture-agnostic method that transforms any pre-trained video diffusion model into an action-conditioned world model, enabling precisely controllable future prediction. Rather than training from scratch, we repurpose the rich world priors of Internet-scale video models by injecting appropriately compressed motor commands through lightweight conditioning layers. This allows our model to follow actions faithfully while preserving generalization and realism. Our approach scales naturally across embodiments and action spaces -- from 3-DoF mobile robots to 25-DoF humanoids, where predicting egocentric joint-angle-driven dynamics is substantially more challenging. The model produces coherent rollouts for both navigation and manipulation, requiring only modest fine-tuning. To evaluate physical correctness independent of appearance, we introduce the Structural Consistency Score (SCS), which measures whether stable scene elements evolve consistently with the provided actions. Our method improves SCS by up to 65\% over the prior state of the art, Navigation World Models; applies seamlessly to three different video diffusion model architectures; and effectively utilizes Internet priors to generalize to unseen environments, including navigation and manipulation inside paintings. Finally, we demonstrate the applicability of EgoWM to robotic planning.
    \keywords{Generative Models \and World Models \and Video Diffusion}
\end{abstract}

\section{Introduction}
\label{sec:intro}

Modeling how the future visual state of the world evolves in response to an agent's actions -- often referred to as \textit{world modeling}~\cite{ha2018world} -- is a fundamental capability for organisms with vision.
It enables planning in navigation and manipulation tasks
and even creative problem-solving~\cite{mcnamee2019internal,pezzulo2024active}.
Animals acquire this ability through a lifetime of interaction
and observation of others, allowing them to anticipate outcomes
even in novel or physically impossible scenarios.

For artificial agents, however, especially for the ones with complex embodiments (like humanoids), action-conditioned videos in diverse real-world settings are expensive to acquire -- an agent must interact with the world to get this data.
As a result, many existing works train world models on narrow,
simulator-defined datasets~\cite{valevski2024diffusion,alonso2024diffusion} or with bespoke designs~\cite{bar2025navigation,zhu2025unified,seo2023masked,wang2024drivedreamer},
which constrains their scalability.
This leads to a crucial question:
can we re-purpose off-the-shelf video models trained on large-scale \textit{passive} data,
and convert them into world models using only a small amount of paired
\textit{action-observation} data?

Several prior works have explored learning action-conditioned
video prediction models~\cite{bar2025navigation, zhu2025irasim, valevski2024diffusion, he2025grndctrlgroundingworldmodels, guo2025ctrl}.
For example, Navigation World Models (NWM)~\cite{bar2025navigation}
trains a 1-billion-parameter diffusion model on egocentric videos to model navigation behavior, while Ctrl-World~\cite{guo2025ctrl} fine-tunes StableVideoDiffusion~\cite{blattmann2023stable} on the large-scale DROID dataset~\cite{khazatsky2024droid} for tabletop manipulation.

However, these methods remain domain-specific, learned for a single robot setup in a limited set of environments. In contrast, modern video generation models~\cite{agarwal2025cosmos,wan2025wan} capitalize on the more scalable Diffusion Transformer (DiT) architecture~\cite{peebles2023scalable} and Internet-scale video-language datasets~\cite{chen2024panda} to enable emulating real-world interactions for any embodiment and environment when given suitable prompts~\cite{wiedemer2025video}. Crucially, these models continue to improve every day~\cite{sora2_openai_2025, seedance2_0_2026}.

The key challenge is to introduce a mechanism for precise controllability, which is universally applicable to \textit{any existing or future} video diffusion model architecture, without compromising generalization acquired during pre-training. We build on the observation that \textit{every} video diffusion architecture
uses a denoising timestep embedding to control the generation process, and inject the action control signal directly into this universal pathway. A central question is aligning actions with the \textit{latent temporal resolution} of each model. We therefore match the model’s latent compression factor and downsample the actions accordingly. The resulting action embeddings are combined with the diffusion timestep embedding, replicated along the temporal dimension of the latent video. 

This design is \textit{architecture-agnostic}: it does not alter the original model's architecture and applies across different backbones.
We validate this with both UNet-based~\cite{ronneberger2015u}
and DiT-based~\cite{peebles2023scalable} models with varying temporal compression factors (Section~\ref{sec:experiments}) and systematically ablate our design choices (Section~\ref{sec:Ablation}). The results show improved action controllability and stronger generalization compared to prior conditioning approaches. Qualitatively, we demonstrate never-before-seen levels of generalization to novel environments, including navigation and manipulation inside paintings (Figure~\ref{fig:paint}; see Section~\ref{sec:qualivid} for discussion).

Our method is also \textit{embodiment-agnostic}. It scales seamlessly from a 3-DoF setting to the 25-DoF joint space of the EVE 1X humanoid, even though large parts of the robot’s body are not visible in the egoview. As illustrated in Figure~\ref{fig:teaser}, this enables high-dimensional humanoid navigation and manipulation -- a capability not demonstrated by prior open-source world models.

To quantitatively evaluate our approach,
we propose a new metric: \textbf{Structural Consistency Score (SCS)} in Section~\ref{sec:scs}.
Prior metrics such as LPIPS~\cite{zhang2018unreasonable} or FVD~\cite{unterthiner2018fvd} focus on perceptual similarity or realism~\cite{liu2024evalcrafter,wickrema2025benchmarking},
which can conflate visual fidelity with physical correctness.
A model could produce sharp, realistic frames that nevertheless depict
physically inconsistent outcomes.
SCS complements these appearance-centric measures by capturing whether the \textit{structure} of the generated environment evolves correctly in response to the agent’s actions, independent of texture. Concretely, we automatically identify \textit{passive} 
(\eg, walls, furniture, objects manipulated by the robot) scene elements, as well as the visible parts of the robot itself, in the first frame and track them across both real and generated videos using a segmentation tracker.
Comparing the resulting trajectories allows us to assess whether the model’s predicted world
evolves coherently with the true environment.
A higher SCS indicates stronger structural and physical consistency
with the provided action sequence. Our approach outperforms prior works by up to 65\% in terms of SCS. Its applicability to downstream robotics tasks is also directly demosntrated in Section~\ref{sec:planning}. \\

\noindent In summary, our contributions are as follows:
\begin{itemize}
    \item We propose a simple yet powerful recipe, \textbf{EgoWM}, to convert \textit{any} image-to-video diffusion model into an
    \textbf{action-conditioned world model}, regardless of base \textit{architecture}, temporal \textit{compression} and \textit{action space} complexity.
    \item We demonstrate that our method works across 
    various video diffusion backbones and extends seamlessly to \textbf{25-DoF} humanoids,
    enabling high-resolution video prediction
    for navigation and manipulation, from an \textbf{egocentric} view, generalizing to highly \textbf{Out-of-Distribution paintings}.
    \item We introduce the \textbf{Structural Consistency Score (SCS)},
    a metric to evaluate the structural and physical plausibility
    of generated videos, measuring action alignment independently, for both \textit{navigation} and \textit{manipulation}.
    \item We demonstrate that EgoWM is effective for downstream robot trajectory \textbf{planning} in both 3-DoF navigation and 25-DoF manipulation.  
\end{itemize}
By learning from large-scale passive datasets and a small amount of action-labeled data, our framework offers a path toward scalable, general-purpose visual dynamics models.

\section{Related Work}
\label{sec:Related_work}
\noindent\textbf{World Models and Video Prediction} have long been studied in reinforcement learning and robotics. The classic work of Ha \etal~\cite{ha2018world} trained a variational RNN to simulate simple games, allowing an agent to plan within its own latent dream. Follow-up works demonstrated video prediction for control in robotics – \eg, visual foresight model~\cite{ebert2018visual}, which learned to predict future camera frames given robot actions and used model-predictive control to push objects to the target location, or SV2P~\cite{babaeizadeh2017stochastic}, which used stochastic latent variables for multi-modal future predictions. These pioneering approaches, however, were limited to relatively constrained environments (toy simulations or single-task robot setups). 

In the gaming domain, several works have proposed to visually emulate a game engine from screen pixels and keyboard actions~\cite{kim2020learning,oh2015action,wang2022predrnn}.
More recently, there is a trend of utilizing high-capacity generative models for world modeling. The DIAMOND agent~\cite{alonso2024diffusion} uses a diffusion-based world model to better preserve visual detail in Atari games. GameNGen~\cite{valevski2024diffusion} takes a similar diffusion-based approach to simulate an entire 3D game in real time. These results underscore the potential of diffusion models to capture complex dynamics, but they have been confined to narrow domains, trained from scratch on domain-specific data.

In another line of work, world models have been trained for tabletop manipulation scenarios~\cite{zhu2025irasim,zhu2025unified,pang2025learning, guo2025ctrl}, but focus on narrow domains with static, exocentric cameras. However, in the wild agents rarely have access to exocentric views and need to move inside their environment. Very recently, Navigation World Model (NWM)~\cite{bar2025navigation} broadened the scope by training a single custom-designed autoregressive diffusion model (cDiT) across embodiments for 3-DoF planar egocentric navigation. However, their design prevents them from leveraging internet-scale video diffusion pretraining, limiting generalisation. In this work, we push further on generality: rather than collecting more data or designing a new model, we adapt existing pre-trained models to a wide variety of egocentric tasks, from 3-DoF navigation to 25-DoF humanoid loco-manipulation.

GrndCtrl~\cite{he2025grndctrlgroundingworldmodels}, IRASim~\cite{zhu2025irasim}, and Ctrl-World~\cite{guo2025ctrl} are recent works that leverage video generation pretraining for navigation and tabletop manipulation.
GrndCtrl re-uses the global action-conditioning mechanism of Cosmos~\cite{agarwal2025cosmos}, which we show to provide limited control precision. IRASim and Ctrl-World build on earlier video diffusion architectures~\cite{blattmann2023stable,zheng2024open} and do not account for temporal compression — a key component of modern DiT-based models.
Our approach, instead, provides a universal conditioning framework compatible with arbitrary video diffusion backbones and enables precise control in complex, high-dimensional action spaces.

\noindent\textbf{Video Diffusion Models}  have emerged as a powerful paradigm for video generation. \cite{ho2022imagen}  demonstrated that a straightforward extension of image diffusion models can produce high-quality short videos.
This was followed by text-to-video models leveraging massive paired visual-language samples~\cite{wang2023modelscope,chen2024videocrafter2,blattmann2023stable}. These models are trained on observation-only video data (\eg, crawled web videos) and are not inherently action-conditioned, but they learn strong priors about physics and appearances. There have also been advances in architectural design: Diffusion Transformers (DiT)~\cite{peebles2023scalable} replaced the U-Net~\cite{ronneberger2015u} backbone, achieving excellent video generation performance~\cite{wan2025wan,agarwal2025cosmos,yang2024cogvideox}.

More recently, diffusion-based generative models have been adapted beyond pure synthesis to support downstream vision and control tasks — leveraging their rich priors to enable generalization to unseen domains. For example, in the image domain, several works~\cite{Liu_2023_ICCV,ozguroglupix2gestalt,zhao2023unleashing,ke2024repurposing} have shown that a pretrained image diffusion backbone can be used to bring several classic vision tasks into the open world. Video diffusion representations have been successfully adapted for video segmentation~\cite{zhu2024exploring,bagchi2025refereverything}, dynamic views synthesis~\cite{bai2025recammaster,van2024generative}, as well as for robot policy learning~\cite{hu2024video,liang2025video}, demonstrating never-before-seen levels of generalization for these tasks. In contrast, in world model learning, most of the approaches -- while adopting the video diffusion objective -- utilize custom model design, limiting the benefits of pre-trained representations.

\noindent\textbf{Evaluation Metrics} used in video generation were originally designed to capture image fidelity or distributional similarity. Frame-level measures such as SSIM~\cite{Wang2004_SSIM} and PSNR~\cite{Mannos1974_PSNR} evaluate brightness, contrast, and structure per frame, while LPIPS~\cite{zhang2018unreasonable} compares neural network features. Sequence-level metrics like Fréchet Video Distance (FVD)~\cite{unterthiner2018fvd}, which measures feature distribution differences using a 3D ConvNet~\cite{Carreira2017_I3D}, were later introduced to capture temporal quality. World modeling approaches typically adopt these metrics to evaluate predicted futures, yet high scores often fail to reflect action alignment. We address this gap with the Structural Consistency Score (SCS), which explicitly disentangles action-following accuracy from visual fidelity.

\section{Method}
\label{sec:method}

\subsection{Preliminaries}
\label{subsec:prelim}

Modern video diffusion models~\cite{blattmann2023stable,chen2024videocrafter2} synthesize future frames conditioned on an observed frame $x_0$ (and optionally text) by iteratively denoising a Gaussian latent. Let $f_{\text{vdm}}$ denote a pre-trained image-to-video model parameterized for $T_s$ diffusion steps. Sampling is performed by drawing $\epsilon \sim \mathcal{N}(0, I)$ and applying the reverse denoising process:
\begin{equation}
    \hat{X} = f_{\text{vdm}}(c_t, \epsilon, T_s),
\end{equation}
where $c_t$ encodes the frame-level conditioning signal.

To reduce computation, these models operate in a low-dimensional latent space. A video $X \in \mathbb{R}^{T \times H \times W \times 3}$ is mapped by a spatio-temporal VAE~\cite{kingma2014autoencoding,rombach2022high} into a latent tensor
$
z = \mathcal{E}(X) \in \mathbb{R}^{\frac{T}{k} \times h \times w \times c},
$
where $k$ denotes the temporal downsampling factor, carefully accounting for which is key for precise action conditioning. 

Generation proceeds by decoding the predicted trajectory:
\begin{equation}
    \hat{X} = \mathcal{D}\!\left(f_{\text{vdm}}(c_t,\epsilon,T_s)\right).
\end{equation}
During training, a noisy latent $z_{t_s}$ is produced by the forward diffusion process, and the model learns to predict the injected noise:
\begin{equation}
\min_\theta 
\mathbb{E}_{z, t_s, \epsilon}\!
\left[
\left\|
\epsilon - \epsilon_\theta(z_{t_s}, e_c, t_s)
\right\|_2^2
\right].
\label{eq:denoise}
\end{equation}
Here $e_c$ denotes the conditioning embedding and $\epsilon_\theta(\cdot)$ is implemented by a U-Net~\cite{ronneberger2015u} or a DiT-based~\cite{peebles2023scalable} denoiser. Although architectures vary widely, all models share a common mechanism: latent features are modulated by the denoising timestep $t_s$, which forms the basis of our action-conditioning strategy.

\subsection{Action Conditioning}
\label{subsec:action}
\begin{figure*}[t]
    \centering
    \includegraphics[width=\linewidth]{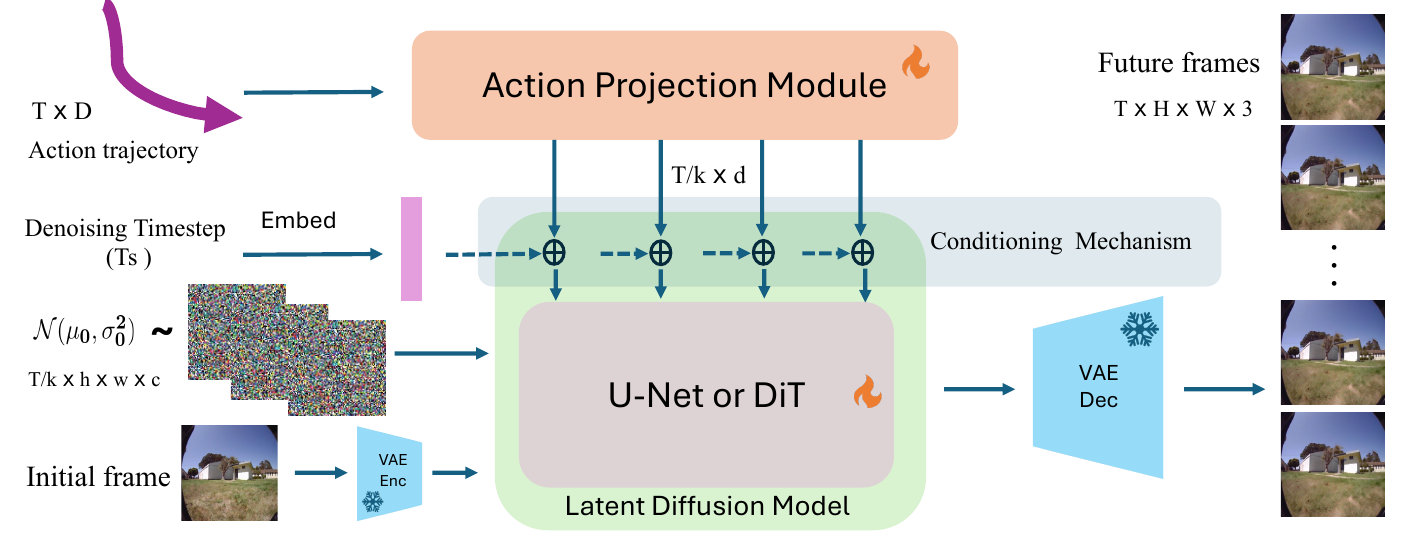}
    \caption{EgoWM embeds high-dimentional action sequences into a universal feature space and injects these embeddings into a pre-trained video diffusion model by reusing its timestep-conditioning pathway (shown in the top center). This enables turning any passive video generation model into a world model without compromising the pre-trained representation.}
    \label{fig:model}
\end{figure*}

Our objective is to convert passive, image-conditioned video diffusion models into action-conditioned world models. Formally, given an action sequence $\mathbf{A} \in \mathbb{R}^{D \times T}$,  where $D$ denotes the action-space dimensionality and $T$ the prediction horizon, and an initial observation $x_0$, the model is fine-tuned to generate future frames:
\begin{equation}
\hat{\mathbf{X}}_{1:T} = f_{\text{vdm}}(x_0, \mathbf{A}, \epsilon, T_s).
\end{equation}
To enable this, we design a general framework, that is architecture-agnostic, allowing a single conditioning strategy to generalize across multiple diffusion models and embodiments. Next, we describe the key components of our approach in more detail.

\noindent\textbf{Action Projection.} 
To condition a video diffusion model on a control trajectory $\mathbf{A}$, we first embed each $D$-dimensional action vector into a latent representation $\mathbf{Z_A} \in \mathbb{R}^{d \times T}$, where $d$ is the embedding dimension. In its general form, the module, shown in Figure~\ref{fig:model} (top), consists of a sequence of lightweight multi-layer perceptrons (MLPs) that map each action vector to the embedding space. For video diffusion models that use temporal latent compression (\ie, VAEs with factor \(k>1\)), action embeddings are further downsampled by 1D convolutions, producing $\mathbf{Z} \in \mathbb{R}^{d \times T/k}$ to ensure alignment with the latent video frame rate.

Different embodiments use different parameterizations: for \emph{3-DoF Navigation}, we adopt the standard control representation $(\Delta x, \Delta y, \Delta\phi)$; for \emph{25-DoF Humanoid Control}, the standard representation is adopted from~\cite{1X_Technologies_1X_World_Model_2024}, but otherwise our design is left unchanged. Because the projection module operates independently of the base architecture and only interfaces through the shared latent space, it can adapt to different temporal resolutions, action dimensionalities, and embodiment types without architectural changes.

\noindent\textbf{Injecting Actions via Timestep Modulation.} Rather than introducing model-specific action conditioning layers, as in prior work~\cite{zhu2025irasim,guo2025ctrl}, we leverage the universal diffusion timestep pathway and adapt it to the model’s latent temporal resolution. Concretely, let the action embeddings be ${Z_a} \in \mathbb{R}^{d \times T/k}$, where $k$ denotes the temporal compression factor. We replicate the global embedding of denoising timestep $t_s$ along the compressed temporal dimension to obtain $Z_{t_s} \in \mathbb{R}^{d \times T/k}$. At each location where the base model applies timestep-dependent modulation to the video latent, we add the corresponding action embedding (Figure~\ref{fig:model}, middle). This enables each temporally aligned action embedding to directly modulate the representation of its associated latent video frames. Formally, for each modulation block $i$, we redefine:
\begin{equation}
P^{\text{scale}}_i, P^{\text{shift}}_i, P^{\text{gate}}_i
= 
F_i\!\left(
Z_{t_s} + Z_a
\right),
\end{equation}
where $F_i(\cdot)$ is the block-specific projection used by the base model.

\noindent\textbf{Initial State Ambiguity in Humanoids.} For humanoid control, where many body parts are not visible in the camera view, we additionally include the embedding of the initial agent state $Z_s$:
\begin{equation}
P^{\text{scale}}_i, P^{\text{shift}}_i, P^{\text{gate}}_i
=
F_i\!\left(
Z_{t_s} + Z_a + Z_s
\right).
\end{equation}

This simple formulation makes the modulation parameters frame-specific and action-aligned, while preserving architectural compatibility with both U-Net–based and DiT-based video diffusion models. Despite its simplicity, this strategy yields fine-grained motion control in both low-DoF and high-DoF settings, while allowing extreme generalization.

\begin{figure*}[t]
    \centering
    \includegraphics[width=\linewidth]{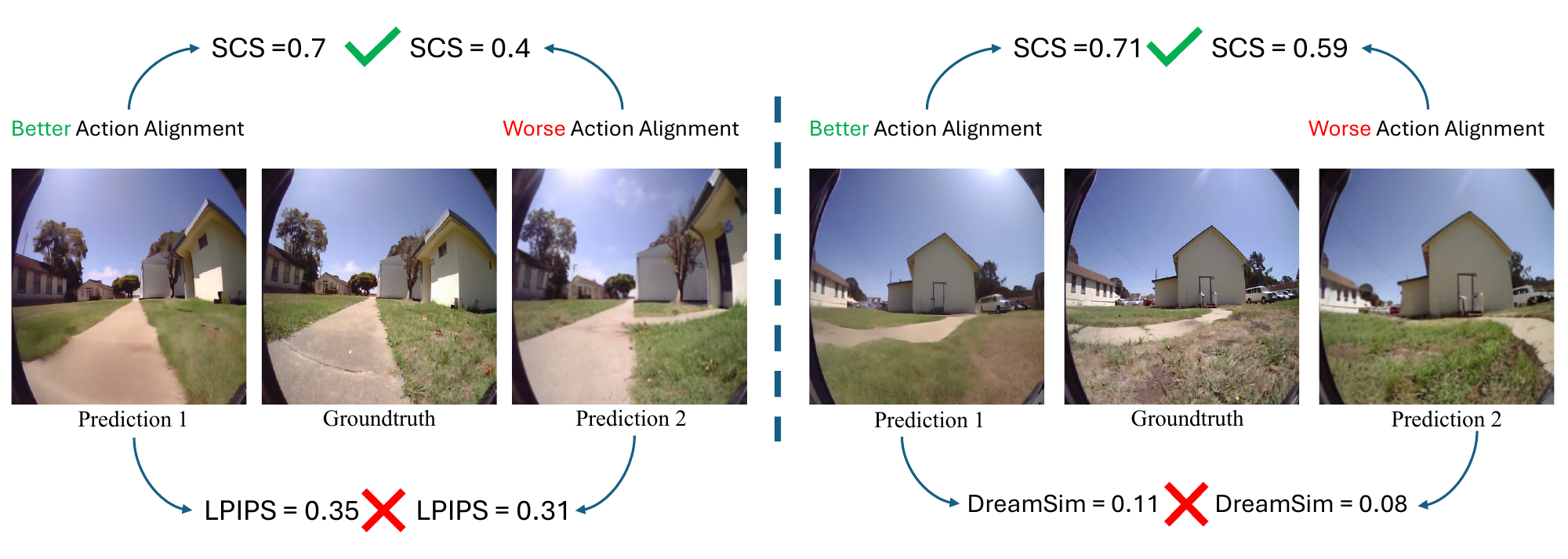}
    \caption{Each column shows two generated rollouts compared to the corresponding ground-truth frame, with their LPIPS, DreamSim, and SCS scores. Perceptual metrics incorrectly favor the visually sharper but physically inconsistent sample, while SCS correctly identifies the sequence that follows the true action trajectory, accurately capturing structural consistency.}
    \label{fig:metric}
\end{figure*}

\section{Structural Consistency Score}
\label{sec:scs}
Existing action-conditioned video generation models are typically evaluated using perceptual similarity metrics such as LPIPS~\cite{zhang2018unreasonable} or DreamSim~\cite{fu2023dreamsim}, computed between the generated and ground-truth frames~\cite{bar2025navigation}.
While effective for measuring visual fidelity, these metrics were designed to capture semantic similarity between object-centric images, not structural alignment in complex scenes. As shown in Figure~\ref{fig:metric}, two model predictions can achieve similar perceptual similarity scores despite very different action-following behavior. Moreover, applying per-frame similarity metrics to long-horizon navigation ignores the inherent stochasticity of the problem: early in a trajectory, a model can meaningfully reconstruct the ground truth, but as the agent moves further, multiple plausible futures emerge.

To address these limitations and accurately evaluate action following, we \textit{augment} the existing appearance-centric metrics with \textbf{S}tructural \textbf{C}onsistency \textbf{S}core (\textbf{SCS}).  
To compute SCS, we first trim each evaluation sequence to exclude frames containing only novel regions unseen in the initial observation, since structural correspondence is ill-defined there. We detect this transition automatically using the dense point tracking method of Harley et al.~\cite{harley2025alltracker}, to identify when all points visible in the first frame have left the field of view.

For the remaining frames, we select key scene structures — passive objects whose apparent motion stems solely from the agent’s actions (\eg, buildings, trees, or objects manipulated by the robot). Restricting evaluation to passive objects reduces ambiguity and allows us to leverage recent foundational models for video object segmentation~\cite{ravi2024sam,carion2025sam}, which, as we show in Section~\ref{sec:Actfollow}, are robust to the artifacts in generated videos. Each selected object is annotated either via sparse point clicks, or via natural language queries (see Section~\ref{sec:Actfollow} for details), and tracked through both ground-truth and predicted sequences. 

The final Structural Consistency Score is obtained by averaging mask IoU~\cite{everingham2010pascal} across all
$T$ frames and $N$ tracked objects:
\begin{equation}
\text{SCS} = \frac{1}{N T}
\sum_{j=1}^{N} \sum_{t=1}^{T}
\frac{|\mathcal{M}_{\text{pred}}^{(t,j)} \cap \mathcal{M}_{\text{gt}}^{(t,j)}|}
     {|\mathcal{M}_{\text{pred}}^{(t,j)} \cup \mathcal{M}_{\text{gt}}^{(t,j)}|},
\end{equation}
where $\mathcal{M}_{\text{pred}}^{(t,j)}$ and $\mathcal{M}_{\text{gt}}^{(t,j)}$
are the binary masks of object $j$ at frame $t$ in the predicted and ground-truth videos, respectively.
Higher SCS values indicate stronger structural alignment between
the predicted and ground-truth sequences.

Figure~\ref{fig:metric} illustrates that SCS correlates strongly with action following, in contrast to perceptual metrics that emphasize visual realism. By focusing on structural evolution rather than appearance, SCS provides a quantitative measure of how faithfully a model predicts the causal consequences of an agent’s actions.

\begin{table*}[t]
    \centering
    \small 
    \caption{Comparison with state-of-the-art methods on the RECON validation set after training on three navigation datasets.
Each column group reports LPIPS, DreamSim, and SCS scores for different prediction horizons. All of our variants, even the temporally compressed (k=4) models, outperform Navigation World Model, especially at longer horizons, highlighting the effectiveness of our approach.}
    \label{tab:allnav}
    \setlength{\tabcolsep}{2pt} 
    \resizebox{\textwidth}{!}{%
    \begin{tabular}{l|ccc|ccc|ccc|ccc}
    \toprule
    Method & 
    \multicolumn{3}{c|}{Frame 2} & 
    \multicolumn{3}{c|}{Frame 4} & 
    \multicolumn{3}{c|}{Frame 8} & 
    \multicolumn{3}{c}{Frame 16} \\ 
    \cline{2-13}
     & LPIPS$\downarrow$ & DreamSim$\downarrow$ & SCS$\uparrow$ & LPIPS$\downarrow$ & DreamSim$\downarrow$ & SCS$\uparrow$ & LPIPS$\downarrow$ & DreamSim$\downarrow$ & SCS$\uparrow$ & LPIPS$\downarrow$ & DreamSim$\downarrow$ & SCS$\uparrow$ \\ 
    \midrule
    \textbf{NWM} $(k=1)$ & 0.26 & 0.10 & 58.4 & 0.30  & 0.11 & 56.2 & 0.35 & 0.13 & 46.8 & 0.45  & 0.19  & 33.4 \\ 
    \textbf{SVD} $(k=1)$ &  0.25 & 0.10  & 62.0 & 0.27 & 0.10 & \textbf{61.7}  & 0.31 & 0.12 & \textbf{57.2} & 0.39 & 0.17  & \textbf{55.2} \\ 
    \textbf{Cosmos-2B} $(k=4)$ & \textbf{0.20} & \textbf{0.07} & \textbf{63.1}& \textbf{0.22} & \textbf{0.08} & 60.4 & \textbf{0.26} & \textbf{0.09} & 56.4 & \textbf{0.33} & \textbf{0.10} & 47.5 \\ 
    \textbf{Wan-14B} $(k=4)$ & 0.22 & 0.08 & 62.7 & 0.26 & \textbf{0.08} & 59.0 & 0.29 & 0.10 & 53.9 & 0.34 & 0.11& 49.7\\ 
    
    \bottomrule
    \end{tabular}
    }

\end{table*}

\section{Experiments}
\label{sec:experiments}
\noindent\textbf{Datasets and Evaluation.}  We evaluate our approach across three progressively challenging settings — (i) 3-DoF ego-centric navigation in real-world robot environments, (ii) humanoid navigation, and (iii) humanoid manipulation with a 25-DoF action space — using four open-source datasets. RECON~\cite{shah2021rapid} contains diverse real-world indoor navigation trajectories. SCAND~\cite{karnan2022socially} features long, continuous paths through structured indoor environments, emphasizing smooth motion and spatial consistency. TartanDrive~\cite{triest2022tartandrive} includes challenging outdoor trajectories with uneven terrain and dynamic backgrounds, testing robustness to visual and proprioceptive noise. Finally, the 1X Humanoid Dataset~\cite{1X_Technologies_1X_World_Model_2024} includes both navigation and manipulation episodes paired with 25-DoF joint-angle states, enabling evaluation of high-dimensional embodied control. 

All datasets are used for training, unless stated otherwise, and we report results on the validation splits of RECON and 1X Humanoid. Following prior work~\cite{bar2025navigation}, we report the standard perceptual similarity metrics LPIPS~\cite{zhang2018unreasonable} and DreamSim~\cite{fu2023dreamsim} to assess visual fidelity between predicted and ground-truth frames. To better capture differences in action following accuracy between the models, we additionally report our SCS metric introduced in Section~\ref{sec:scs}. We report results up to 4 seconds into the future at 4 FPS, with additional results over 4$\times$ longer horizons provided in Section~\ref{sec:long_horizon_supp}.

\noindent\textbf{Implementation Details.} 
Across all experiments, we evaluate three variants of our method built on top of the SVD~\cite{blattmann2023stable}, Cosmos~\cite{agarwal2025cosmos} (Predict2-2B) and Wan2.1-14B~\cite{wan2025wan} models. We use two 1D convolutional layers to downsample the action embeddings to match the temporally compressed latent resolution of Cosmos and Wan $(k=4)$. Fine-tuning is performed using the original pretraining objective of the base diffusion model (denoising or flow matching; see Section~\ref{subsec:prelim}). The learning rate for the newly introduced action projection module is set to 10$\times$ higher than that of the base model weights.

\begin{figure*}[t]
    \centering
    \includegraphics[width=\linewidth]{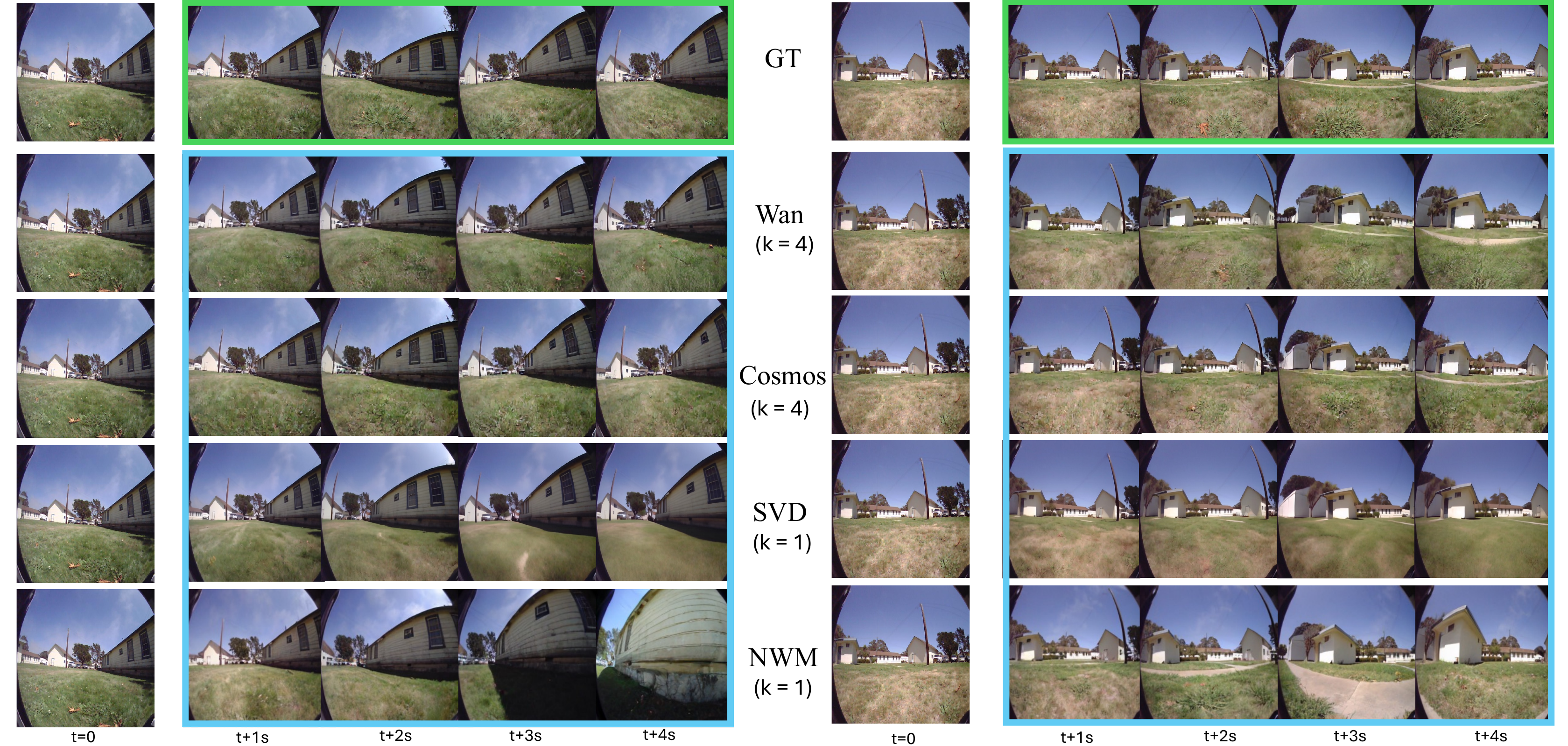}
    \caption{Qualitative results on egocentric navigation.
All variants of our model generate realistic, temporally coherent sequences that accurately follow the provided action trajectories regardless of compression, while NWM exhibits noticeable drift (\eg, veering right in the left column). Results best viewed on the \href{https://egowm.github.io}{project page}.}
    \label{fig:allnav}
\end{figure*}

\subsection{Navigation Results}
We fine-tune all three (SVD, Cosmos, and Wan) variants of our method on the training sets of RECON, SCAND, and Tartan-Drive for a fair comparison with NWM~\cite{bar2025navigation} and report results on the validation set of RECON. Table~\ref{tab:allnav} reports LPIPS, DreamSim, and our proposed SCS metrics for prediction horizons of 2 to 16 frames. All variants of our method outperform NWM across all evaluation horizons and metrics, with particularly large improvements in SCS and at greater distances from the initial frame (up to +65\%), indicating substantially stronger action alignment. Cosmos and Wan achieve higher perceptual fidelity, reflecting the stronger visual priors induced by large-scale pretraining, while SVD exhibits more stable structural consistency over long trajectories due to the absence of temporal downsampling. Nevertheless, our action conditioning design enables modern, temporally compressed DiT-based models to attain strong SCS performance despite operating at reduced latent temporal resolution.

Figure~\ref{fig:allnav} visualizes predicted rollouts on the RECON validation set. All variants of our model produce realistic sequences that closely follow the provided trajectories, while NWM exhibits noticeable drift. Cosmos and Wan generate sharper and more detailed predictions, whereas SVD exhibits mild fidelity loss due to its chunk-wise autoregressive inference. Overall, our results demonstrate that leveraging pre-trained video diffusion models for world modeling yields substantial improvements over specialized NWM architecture, combining high perceptual quality with robust action following. 

In terms of efficiency, NWM is trained on 64 H100 GPUs and predicts at $224\times224$, while our SVD and Cosmos variants use only 8 A100 GPUs and generate higher-resolution outputs ($512\times512$ and $480\times640$, respectively). On the same 20 RECON samples using a single A100 GPU, EgoWM achieves up to $6\times$ faster 64-frame inference than NWM, while using less action data and approximately $8\times$ less training compute; see Section~\ref{sec:compute_latency_supp} for the detailed compute, data, resolution, and latency comparison.

\begin{table*}[t]
    \centering
    \small
    \caption{Results of SVD, Cosmos, Wan and ablation variants on humanoid navigation and manipulation. Our approach allows seamless generalization to this complex 25-DoF action space. The pre-trained SVD variant outperforms the baseline trained from scratch by significant margins, demonstrating the value of Internet-scale pre-training in this challenging scenario.}
    \label{tab:1x}
    \setlength{\tabcolsep}{1pt}
    \resizebox{\textwidth}{!}{%
    \begin{tabular}{c|l|ccc|ccc|ccc|ccc}
    \toprule
    Task & Variant & 
    \multicolumn{3}{c|}{Frame 2} & 
    \multicolumn{3}{c|}{Frame 4} & 
    \multicolumn{3}{c|}{Frame 8} & 
    \multicolumn{3}{c}{Frame 16} \\ 
    \cline{3-14}
      &  & LPIPS$\downarrow$ & DreamSim$\downarrow$ & SCS$\uparrow$ 
      & LPIPS$\downarrow$ & DreamSim$\downarrow$ & SCS$\uparrow$ 
      & LPIPS$\downarrow$ & DreamSim$\downarrow$ & SCS$\uparrow$ 
      & LPIPS$\downarrow$ & DreamSim$\downarrow$ & SCS$\uparrow$ \\ 
    \midrule
    \multirow{3}{*}{Navigation} 
        & \textbf{SVD} & 0.11 & 0.07 & \textbf{75.6} & 0.16 & 0.09 & \textbf{66.4} & \textbf{0.26} & 0.13 & \textbf{50.3} & \textbf{0.35} & 0.18 & \textbf{34.4} \\ 
        & \textbf{SVD} (scratch) & 0.17 & 0.15 & 69.7 & 0.25 & 0.20 & 56.0 & 0.37 & 0.30 & 40.0 & 0.47 & 0.40 & 21.6 \\ 
        & \textbf{Cosmos-2B} & 0.11 & 0.04 & 65.2 & 0.19 & 0.06 & 54.0 & 0.27 & 0.09 & 42.3 & 0.40 & 0.19 & 27.0 \\ 
        
        & \textbf{Wan-14B} & \textbf{0.09} & \textbf{0.04} & 70.2 & \textbf{0.15} & \textbf{0.06} & 61.3 & 0.27 & \textbf{0.10} & 45.2 & 0.39 & \textbf{0.17}  & 30.2 \\ 
    \midrule
    \multirow{2}{*}{Manipulation} 
        & \textbf{SVD} & 0.04 & 0.04 & \textbf{86.8} & \textbf{0.06} & 0.05 & 81.8 & 0.10 & 0.07 & 75.0 & 0.13 & 0.10 & \textbf{76.6} \\ 
        & \textbf{Cosmos-2B} & \textbf{0.04} & \textbf{0.03} & 86.2 & 0.07 & \textbf{0.05} & \textbf{82.0} & \textbf{0.10} & \textbf{0.06} & \textbf{78.2} & \textbf{0.12} & \textbf{0.07} & 67.7 \\ 
        & \textbf{Wan-14B} & 0.05 & 0.04 & 81.1 & 0.07 & 0.06 & 78.5 & 0.11 & 0.07 & 76.0 & 0.14 & 0.09 & 65.6\\ 
    \bottomrule
    \end{tabular}
    }
\end{table*}

\subsection{Humanoid Results}
\begin{figure*}[t]
    \centering

    \includegraphics[width=\linewidth]{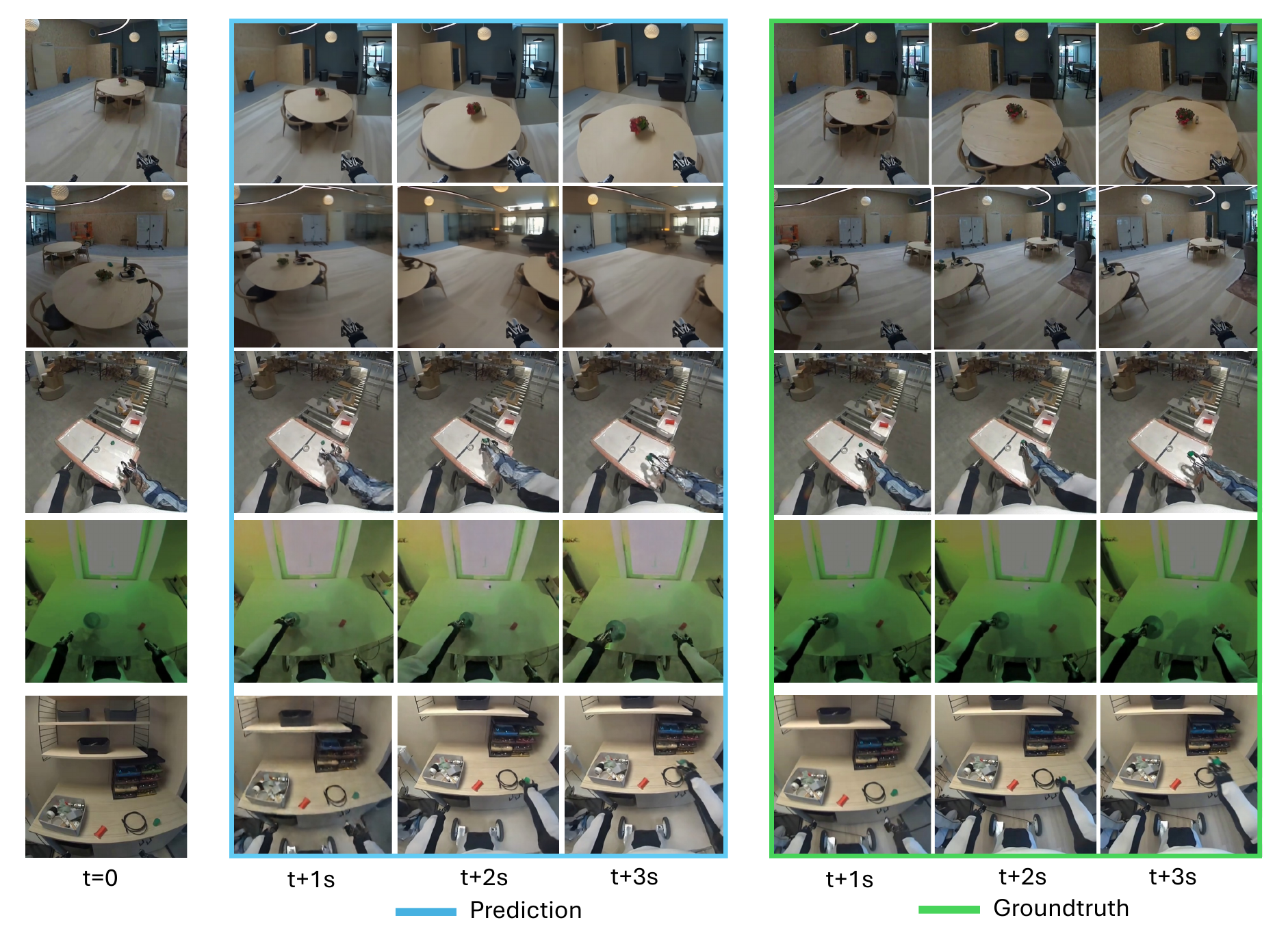}
    
    \caption{Our action-conditioned video world model enables a humanoid agent to navigate and execute long-horizon, contact-rich manipulation skills by conditioning on desired actions. Across diverse scenes and object geometries, the model produces physically consistent body motions, stable grasps, and smooth end-effector trajectories that closely follow the provided action sequence. Results best viewed on the \href{https://egowm.github.io}{project page}.}
    \label{fig:1xmanip}
\end{figure*}

We next evaluate our approach on the 1X Humanoid Dataset, a significantly more challenging setting involving a 25-DoF control space spanning all major joints, neck motion, and gripper actions. As shown in the 25-DoF navigation results in Table~\ref{tab:1x}, all variants exhibit somewhat lower SCS scores toward the end of the trajectory compared to the 3-DoF navigation experiments. This reflects the increased difficulty of high-dimensional humanoid control, where maintaining coherent motion requires reasoning over many coupled joints and partial ego-centric visibility. Nevertheless, our universal conditioning approach applies to both embodiments without any architectural modifications, demonstrating its scalability across action spaces that differ by nearly an order of magnitude in complexity. The SVD model also outperforms its variant trained from scratch, with the gap being especially significant on perceptual similarity, confirming the value of Internet-scale, passive pretraining. We further compare to concurrently proposed humanoid WM architectures~\cite{gao2026dreamdojo} in the ablation analysis in Section~\ref{sec:Ablation}.

In the lower part of Table~\ref{tab:1x}, we quantitatively evaluate our models on 25-DoF manipulation. Since the scene changes very little during manipulation compared to navigation, the LPIPS and DreamSim scores are much lower even at frame 16, compared to navigation. We also see high SCS scores, showing that all variants learn to accurately follow the manipulation commands. SVD is better at action following further into the future, compared to Cosmos and Wan, a common trend across all results, as a consequence of the temporal compression.

Figure~\ref{fig:1xmanip} illustrates the prediction of our model on humanoid sequences in the validation set of 1X. Despite the dramatic increase in embodiment complexity to 25 DoF, EgoWM produces stable trajectories that remain consistent with the provided actions for navigation, reaching, and grasping. Our model maintains coherent global scene structure while capturing detailed arm, hand, and gripper articulation throughout. Crucially, the predicted videos remain visually consistent and physically plausible, enabling planning applications in Section~\ref{sec:planning}.

\subsection{Ablation Analysis}
\label{sec:Ablation}
In Table~\ref{tab:ablations}, we compare the core design choices of EgoWM against alternative conditioning strategies used in prior work. We first assess the effectiveness of our universal action-conditioning mechanism in the high-dimensional humanoid action space by comparing it to two common alternatives: global conditioning used in~\cite{he2025grndctrlgroundingworldmodels} and the Cosmos-Predict-2B World Model [\href{https://huggingface.co/nvidia/Cosmos-Predict2-2B-Sample-Action-Conditioned}{huggingface}], and the chunking strategy proposed in concurrent work~\cite{gao2026dreamdojo}. While perceptual metrics remain largely unchanged across designs, controllability, measured by SCS, drops by 44.8\% under global conditioning and by 12\% with chunking. These results highlight the importance of temporally aligned action injection for precise control.
\begin{table*}[t]
    \centering
    \small
    \caption{Ablation studies at the 16-frame horizon. We first validate the accuracy of our action-injection mechanism on humanoid navigation, and then report out-of-distribution evaluation on RECON to assess generalization. Our design achieves both more accurate controllability and stronger generalization compared to the alternatives.}
    \label{tab:ablations}
    \setlength{\tabcolsep}{4pt}
    \resizebox{\textwidth}{!}{%
    \begin{tabular}{l|l|ccc}
    \toprule
    Dataset / Task & Variant & LPIPS$\downarrow$ & DreamSim$\downarrow$ & SCS$\uparrow$ \\
    \midrule
    \multicolumn{5}{l}{\textbf{Complex action space controllability at $k = 4$}} \\
    \midrule
    Humanoid Navigation
        & Cosmos-2B (Ours) & \textbf{0.40} & 0.19 & \textbf{27.0} \\
         & Cosmos-2B (Global) & 0.46 & 0.20 & 14.9 \\
        & Cosmos-2B (Chunked) & 0.42 & \textbf{0.17} & 23.7 \\
    \midrule
    \multicolumn{5}{l}{\textbf{Out-of-distribution generalization}} \\
    \midrule
    RECON (OOD) 
        &  SVD (Ours) & \textbf{0.59} & \textbf{0.43} & \textbf{29.7} \\
        & SVD (X-attn) & 0.64 & 0.46 & 22.7 \\
    \bottomrule
    \end{tabular}
    }
\end{table*}
Next, we evaluate how the action-conditioning pathway affects generalization by comparing our method to the cross-attention conditioning used in the recent Ctrl-World model~\cite{guo2025ctrl}. As out-of-distribution data is not available for the 1X humanoid embodiment, we conduct this study in the 3-DoF navigation setting using SVD as the base model to ensure a faithful reproduction of~\cite{guo2025ctrl}. Both variants are trained on SCAND (\textit{urban campus}) and evaluated on the unseen RECON environments (\textit{residential suburbs}). We observe that our denoising timestep-based conditioning is better at preserving priors learned during Internet-scale pre-training of video diffusion models, compared to the cross-attention approach of Ctrl-World, as indicated by the appearance-centric metrics. Moreover, it also provides more precise action controllability, yielding a 23.6\% gain in SCS.

\subsection{Extreme Generalization}
\begin{figure}[t]
    \centering
    \includegraphics[width=0.8\linewidth]{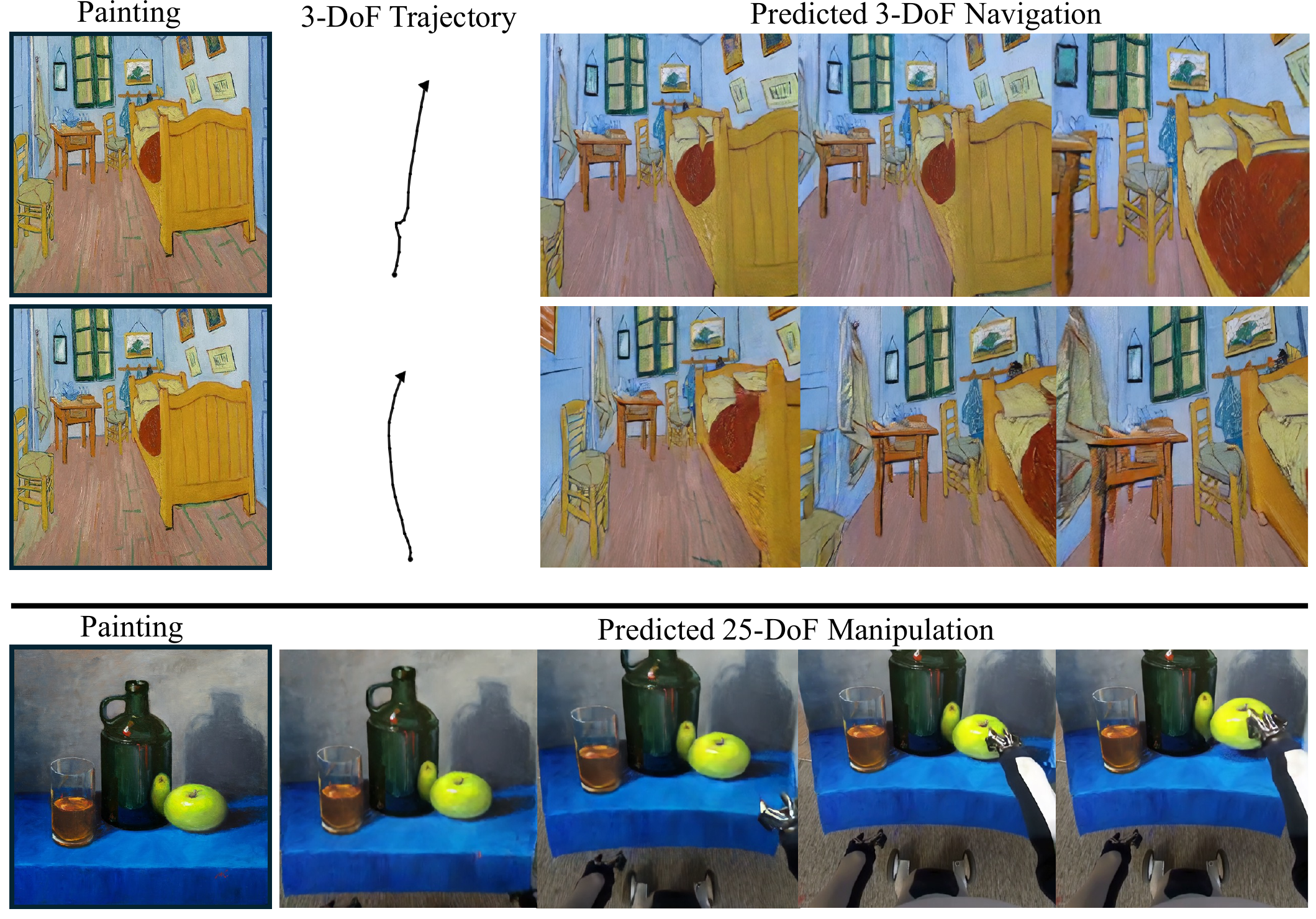}
    \caption{EgoWM generalizes to paintings, including navigating inside a room and picking up an apple from a table. Despite the drastic shift in appearance, EgoWM preserves coherent motion, effectively ``walking into'' and ``interacting with'' painted worlds by leveraging Internet-scale pre-training. Results best viewed on the \href{https://egowm.github.io}{project page}.}
    \label{fig:paint}
\end{figure}

Figure~\ref{fig:paint} shows EgoWM generalization results in highly unrealistic domains. In the top part of the figure, we evaluate our 3-DoF Wan variant by conditioning on the same initial painting and generating distinct, unseen navigation trajectories from it, highlighting its ability to produce different motion-consistent futures in OOD scenarios. We further show similar generalization in 25-DoF manipulation, where the Wan variant rolls out an action sequence from the 1X validation set to simulate a humanoid looking down, reaching with the right arm, and picking up an apple inside a painting. More results are available on the \href{https://egowm.github.io}{project page}.

\section{Planning with EgoWM}
\label{sec:planning}
In this section, we evaluate whether EgoWM can support downstream decision-making with its simulated futures, allowing model-based action selection.

\noindent\textbf{Experiment Setup.}
We use goal-conditioned planning with the Cross-Entropy Method (CEM), following NWM~\cite{bar2025navigation}. Candidate action sequences are sampled, rolled out with our world model, and selected based on perceptual similarity between their predicted final frames and a goal image. 

We evaluate this setup in two settings: 3-DoF navigation on RECON and 25-DoF humanoid manipulation on held-out 1X validation tasks. For navigation, we follow the NWM protocol and use the same trajectory prior: candidate trajectories are straight-line motions with equidistant intermediate steps. We sample 120 trajectories radiating from the robot's current pose, roll each candidate out in EgoWM, compare the predicted final frame to the goal image, and average the top-scoring actions to obtain the final planned trajectory. It is then evaluated against the ground-truth trajectory using Average Trajectory Error (ATE)~\cite{bar2025navigation}.

For manipulation, as there are many possible ways to successfully complete a task, we measure the success rate of the generated trajectories (SR) instead of ATE. A simulator is not available for 1X, so we use human assessment to compute SR, following Ctrl-World~\cite{guo2025ctrl}. To sample  candidate trajectories, we split episodes into validation and test subsets and use the val set to estimate CEM sampling parameters. We then apply the inference procedure described above on the test set to obtain the final planned trajectories. Only held-out tasks that were not used to train EgoWM are included in the evaluation to test generalization.

\begin{table}[t]
\centering
\scriptsize
\setlength{\abovecaptionskip}{2pt}
\setlength{\belowcaptionskip}{2pt}
\begin{minipage}[t]{0.42\linewidth}
\centering
\caption{3-DoF navigation planning evaluation on RECON over 16-frame trajectories (4 seconds). Both EgoWM variants achieve lower average trajectory error (ATE).}
\label{tab:3dofplanning_main}
\setlength{\tabcolsep}{2pt}
\begin{tabular}{lc}
\toprule
Method & ATE $\downarrow$ \\
\midrule
NWM & 3.25 \\
EgoWM (SVD) & 3.18 \\
EgoWM (Cosmos) & \textbf{3.13} \\
\bottomrule
\end{tabular}
\end{minipage}
\hfill
\begin{minipage}[t]{0.55\linewidth}
\centering
\caption{25-DoF manipulation results on 1X validation tasks. We report success rate, with planned trajectories selected by EgoWM and GT demonstrations evaluated in EgoWM.}
\label{tab:manip_planning_main}
\setlength{\tabcolsep}{2pt}
\resizebox{\linewidth}{!}{%
\begin{tabular}{lcc}
\toprule
Task & SR (Planned) $\uparrow$ & SR (GT) $\uparrow$ \\
\midrule
P\&P towel in basket & 100.0\% & 100.0\% \\
P\&P block on board & 80.0\% & 94.0\% \\
Pick up obj from pile & 60.0\% & 75.0\% \\
Pick up cup & 66.7\% & 100.0\% \\
\bottomrule
\end{tabular}}
\end{minipage}
\end{table}

\noindent\textbf{Results.}
In 3-DoF navigation, as shown in Table~\ref{tab:3dofplanning_main}, both EgoWM variants outperform NWM. The gains are moderate because this short-horizon setting is strongly constrained by the straight-line trajectory prior and dense coverage of sampled directions. Still, the improvement indicates that EgoWM predictions can effectively support navigation planning.

In 25-DoF humanoid manipulation, we report results with the Cosmos backbone in Table~\ref{tab:manip_planning_main} (middle) and observe that CEM-selected trajectories achieve high success across several tasks, despite the higher-dimensional action space. We further report the success rate of ground-truth demonstrations rolled out in EgoWM in the third column. For most tasks, the target success rate of 100\% is recovered, demonstrating the potential of our model for policy evaluation.

\section{Conclusion}
\label{sec:conclusion}
We presented a simple and general framework for transforming pre-trained video diffusion models into world models. By introducing lightweight action conditioning into existing architectures, our method leverages the priors learned from Internet-scale data to produce physically and semantically consistent futures. Through extensive experiments, we demonstrated strong generalization across diverse embodiments and environments, including humanoids, and effectiveness in downstream decision-making. Finally, we proposed the Structural Consistency Score (SCS) to faithfully evaluate world models' action following. These contributions open a path toward scalable, controllable, and generalizable world models, bridging the gap between passive and active visual prediction.

\subsection*{Acknowledgment} This project was supported by Toyota Research Institute.

\bibliographystyle{splncs04}
\bibliography{main}

\clearpage
\appendix
\begin{center}
{\Large\bfseries Appendix}
\end{center}
\setcounter{section}{0}
\renewcommand{\thesection}{\Alph{section}}
\renewcommand{\theHsection}{appendix.\Alph{section}}
In this appendix, we report additional results, details, and visualizations that were not included in the main paper due to space limitations. We begin by discussing our qualitative video results provided on the \href{https://egowm.github.io}{webpage} in Section~\ref{sec:qualivid}. Next, we discuss the Structural Consistency Metric in Section~\ref{sec:Actfollow}. We then report further comparisons to the Navigation World Models (NWM)~\cite{bar2025navigation} baseline in Section~\ref{sec:nwmcmp} and provide additional implementation details in Section~\ref{sec:impl_supp}. Finally, we discuss failure modes of our method in Section~\ref{sec:failure}.

\section{Qualitative Video Results}
\label{sec:qualivid}
This section provides an overview of the \href{https://egowm.github.io}{webpage}, which presents our model predictions and comparisons with the NWM baseline in video format.

\noindent\textbf{Zero-shot Generalization to Paintings.}
We first provide qualitative video examples of zero-shot 25-DoF manipulation and 3-DoF navigation in paintings. To our knowledge, this is the first work to demonstrate humanoid manipulation of objects depicted in paintings. In each block, we show a single painting together with two distinct action trajectories simulated from the same initial scene (the original humanoids videos which were used to source the trajectories are also shown for reference). Our world model generates diverse behaviors, such as picking up an apple or pulling a tablecloth, accurately following the action sequences. While some examples exhibit physically inaccurate grasps, this is expected given that object interactions in painted domains are absent from the training data. We additionally demonstrate 3-DoF position control, where the model accurately follows both position and velocity of the specified trajectories.

\noindent\textbf{Zero-shot Generalization to Real-World Scenes.}
While paintings illustrate the extremes of our model’s generalization, a more practical setting involves unseen real-world environments. We therefore present 25-DoF humanoid navigation in real-world scenes that were not observed during training. The model generalizes to these environments and follows the ground-truth navigation actions shown to the left of each block.

\noindent\textbf{In-Domain Comparisons.}
Here we show comparisons between different variants of our model (SVD, Cosmos, Wan), and against NWM for 3-DoF navigation. On 3-DoF navigation in the RECON test set, NWM exhibits substantial drift over time, whereas all variants of our method ($k \geq 1$) remain consistent with the commanded actions. Among the base models, SVD preserves fewer scene details compared to the others. For 25-DoF manipulation on the 1X validation set, all variants of our model demonstrate controlled humanoid behaviors, including looking down, reaching, and picking up objects.

\noindent\textbf{Fine-grained Humanoid Control.}
We demonstrate the expressiveness of the 25-DoF action space by simulating action sequences that differ only subtly from similar initial states, for both navigation and manipulation on the 1X validation set. All results in this section are generated using Wan and Cosmos ($k=4$), both temporally compressed models. These examples illustrate that our control mechanism enables consistent and distinguishable behaviors even under small variations in the commanded actions. We also show two failure cases in re-arranging small blocks on a table, with our models struggling with object permanence. We discuss failure modes in detail in Section \ref{sec:failure}.


\section{Structural Consistency Evaluation}
\label{sec:Actfollow}
In the main paper, in order to segment key structures in the scene 
we used SAM2's point annotation interface. Specifically, we manually selected key points in the initial frame corresponding to salient structural elements that define the scene layout (e.g., houses, trees, the robot arm). SAM2 was then used to tracks these entities, enabling computation of the metric introduced in Section 4. In this section, we visually illustrate the intuition behind the metric, validate its robustness to distortions in generated videos, and show how the manual point-annotation step can be removed to scale the metric to large datasets without sacrificing accuracy.

\begin{figure*}[t]
    \centering
    \includegraphics[width=0.75\linewidth]{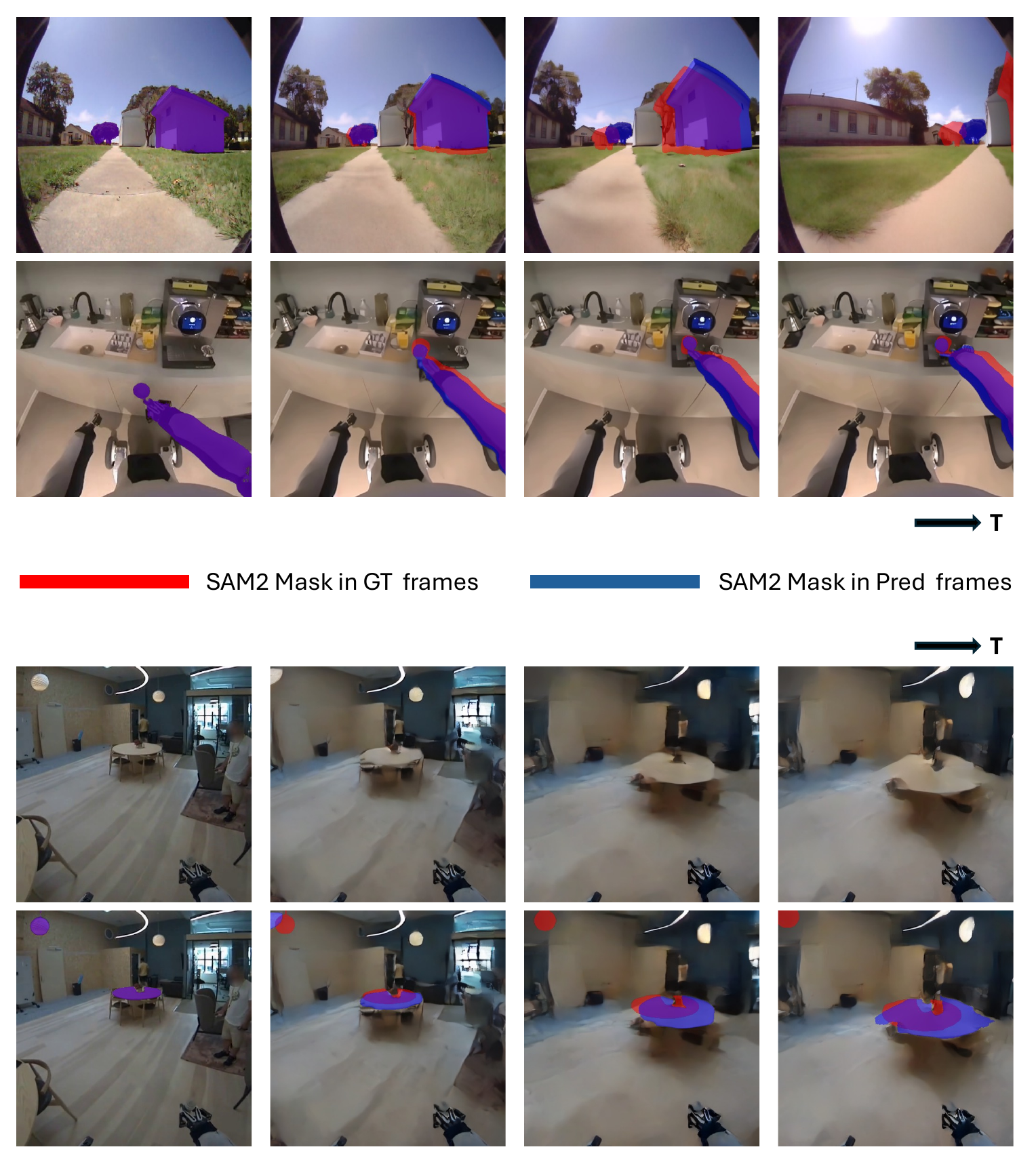}
    \caption{Illustration of our SCS metric computation. The first two rows show the difference between predicted locations of structures and their groundtruth location for navigation and manipulation. The last two rows demonstrate the robustness of our metric to distorted generations, with the third row showing the distorted predicted frames and the final row showing the tracked masks of these distorted structures and their groundtruth positions.}
    \label{fig:scs_illus}
\end{figure*}
\noindent\textbf{SCS Illustration.} In Figure~\ref{fig:scs_illus} we qualitatively demonstrate the structural consistency evaluation using our SCS metric, for navigation and manipulation. The first row shows predicted frames for 3-DoF navigation in RECON, with two key structures (a house and a tree) tracked with blue masks, and the positions of these structures in the corresponding ground-truth frames shown with red masks. As we can see, initially the two structures' predicted location and shape is perfectly aligned with ground-truth but, as we go further into the future, the imperfections in action following result in a drift. Second row shows an evaluation for 25-DoF manipulation using humanoid. Note here that while the hand's predicted motion is very close to the ground-truth, the shape of the cup gets distorted during manipulation, and will be correctly penalized by our SCS metric.

\noindent\textbf{SCS Applicability for Long-horizon Evaluation.}
The metric evaluating only objects that remain visible in the frame is by design, not a limitation. Predicting entirely new content entering the scene is fundamentally ill-defined; the model cannot accurately anticipate what objects will appear. That said, as shown in Table~\ref{tab:scs_early_termination}, only 17\% of tracked segments leave the view by frame 64 in navigation evaluation, so SCS remains valid across our tested horizons.

\begin{table}[t]
\centering
\small
\caption{Fraction of tracked segments that leave the view during long-horizon navigation evaluation.}
\label{tab:scs_early_termination}
\setlength{\tabcolsep}{8pt}
\begin{tabular}{c|c}
\toprule
Frame & Early Termination \\
\midrule
8 & 0\% \\
16 & 0\% \\
32 & 2\% \\
64 & 17\% \\
\bottomrule
\end{tabular}
\end{table}

\noindent\textbf{Robustness to Distortions.}
SCS is designed to measure structural alignment rather than rendering fidelity. In the bottom two rows of Figure~\ref{fig:scs_illus}, we illustrate this property on highly distorted 25-DoF navigation predictions from an intermediate stage of model training. The final row shows the corresponding SAM2 segmentation results for the light fixture and table in blue, with ground-truth object segmentation overlaid in red. Despite severe visual artifacts, SAM2 mask tracking remains stable, allowing SCS to evaluate action following even when generation fidelity is poor. This qualitative behavior is consistent with the main-paper observation that models with lower rendering quality can achieve higher SCS (SVD vs. Wan), highlighting the intended disentanglement between action following and perceptual quality.

We further quantify this robustness by adding varying amounts of Gaussian noise to frames using the diffusion schedule and evaluating SAM2 accuracy via mIoU. As shown in Table~\ref{tab:scs_noise_robustness}, tracking remains accurate even under heavy distortion: at PSNR 16.5, SAM2 still achieves 0.93 mIoU. These results demonstrate that SCS probes genuine structural consistency rather than sensitivity to superficial rendering artifacts.

\begin{table}[t]
\centering
\small
\caption{SAM2 tracking robustness under increasing frame distortion.}
\label{tab:scs_noise_robustness}
\setlength{\tabcolsep}{6pt}
\begin{tabular}{l|c|c}
\toprule
Noise & PSNR & mIoU \\
\midrule
10 (mild) & 33.4 & 0.98 \\
50 (noticeable) & 21.8 & 0.96 \\
100 (heavy) & 16.5 & 0.93 \\
\bottomrule
\end{tabular}
\end{table}

\begin{table}[t]
\centering
\small
\caption{Comparison of manual vs automatic variants of SCS at different horizons. Automatic version of our metric preserve the relative order of the methods across time horizons, removing the need for manual annotation of test sequences.}
\label{tab:auto_eval}
\setlength{\tabcolsep}{4pt}

\begin{tabular}{l|cc|cc}
\toprule
\textbf{Method} & \multicolumn{2}{c|}{SCS@1s $\uparrow$} & \multicolumn{2}{c}{SCS@4s $\uparrow$} \\
\cline{2-5}
 & Manual & Auto & Manual & Auto \\
\midrule
NWM & 56.2 & 56.6 & 33.4 & 36.3 \\
SVD & 61.7 & 61.9 & 55.2 & 52.2 \\
Cosmos-2B & 60.4 & 60.2 & 47.5 & 49.1 \\

\bottomrule
\end{tabular}

\end{table}

\noindent\textbf{SCS Automation.}
Although the results reported in the main paper rely on manually selected key structures, the metric itself does not require manual point annotation. To demonstrate this, we evaluate a fully automated variant using SAM3 for 3-DoF navigation on RECON. Instead of manually specifying key points, we provide a predefined set of semantic category prompts corresponding to dominant structural elements in the environment (e.g., trees, houses). Given these text prompts, SAM3 segments and tracks all matching instances in both predicted and ground-truth videos. 

Results are shown in Table~\ref{tab:auto_eval}. Automating the segmentation and tracking stage preserves the relative ranking of methods across the prediction horizons. This indicates that the metric can be scaled to large datasets without manual intervention, to precisely measure structural consistency independent of visual fidelity. 

\section{Long-Horizon Prediction}
\label{sec:long_horizon_supp}
To assess long-horizon modeling capacity, we evaluate 64-frame autoregressive rollouts, which are $4\times$ longer than the 16-frame horizon reported in the main quantitative tables. We compute SCS for both 3-DoF navigation and 25-DoF manipulation on the subset of validation trajectories that are at least 64 frames long.

\begin{table}[t]
\centering
\small
\caption{Long-horizon SCS scores for 64-frame autoregressive rollouts. EgoWM maintains strong action controllability at horizons substantially longer than those in the main quantitative tables.}
\label{tab:long_horizon_scs_supp}
\setlength{\tabcolsep}{4pt}
\begin{tabular}{l|l|c|c|c|c}
\toprule
Task & Method & @8F & @16F & @32F & @64F \\
\midrule
Nav. & NWM & 0.47 & 0.33 & 0.17 & 0.15 \\
& SVD & 0.57 & 0.55 & 0.50 & 0.41 \\
& Wan & 0.54 & 0.50 & 0.40 & 0.33 \\
\midrule
Manip. & SVD & 0.70 & 0.67 & 0.69 & 0.61 \\
& Wan & 0.72 & 0.73 & 0.70 & 0.62 \\
\bottomrule
\end{tabular}
\end{table}

As shown in Table~\ref{tab:long_horizon_scs_supp}, EgoWM maintains strong action controllability over long rollouts. In navigation, both SVD and DiT variants degrade less significantly than the autoregressive NWM baseline. In manipulation, SCS remains largely stable through 64 frames. These results demonstrate that EgoWM scales to horizons relevant for policy learning.

\section{Compute and Latency}
\label{sec:nwmcmp}
\label{sec:compute_latency_supp}
Table~\ref{tab:method_datasets_compute} lists the action datasets, training compute, output resolution, and inference latency for NWM and our 3-DoF EgoWM variants. NWM uses a higher-resolution version of the Huron dataset, which we did not have access to, and is trained on 64 H100 GPUs. In contrast, our SVD and Cosmos variants are trained on RECON, SCAND, and Tartan-Drive using only 8 A100 GPUs.

For evaluation fairness, all reported metrics in the main paper are computed at $224\times224$ by downsampling our predictions to match NWM, even though our models generate at substantially higher native resolutions: $512\times512$ for SVD and $480\times640$ for Cosmos. Thus, EgoWM improves over NWM while using less action data, approximately $8\times$ less training compute, and higher-resolution prediction.

To compare inference latency, we run each model on the same 20 RECON samples across varying prediction horizons using a single A100 GPU and each model's native resolution: $224\times224$ for NWM, $512\times512$ for EgoWM (SVD), and $480\times640$ for EgoWM (Cosmos). EgoWM achieves up to \textbf{$6\times$ faster} 64-frame inference while predicting at up to \textbf{$2\times$ higher resolution}. Cosmos is fastest because it predicts longer chunks in a single forward pass, whereas SVD requires chunk-wise autoregressive inference.

\begin{table*}[t]
    \centering
    \small
    \caption{Comparison of data, compute, native resolution, and inference latency with NWM for 3-DoF position control. All metric comparisons in the main paper are computed at $224\times224$ by downsampling our predictions to match NWM.}
    \label{tab:method_datasets_compute}
    \setlength{\tabcolsep}{3pt}
    \resizebox{\textwidth}{!}{%
    \begin{tabular}{l|c|c|c|c}
    \hline
    \textbf{Method} & \textbf{3-DoF Action Data} & \textbf{Compute} & \textbf{Native Resolution} & \textbf{Avg. 64f Latency (s)} \\
    \hline
    NWM & RECON, SCAND, Tartan-Drive, Huron & 64$\times$ H100s & 224$\times$224 & $\approx 300$ \\
    \hline
    EgoWM (SVD) &
    \multirow{2}{*}{RECON, SCAND, Tartan-Drive} &
    \multirow{2}{*}{8$\times$ A100s} &
    512$\times$512 & $\approx 200$ \\
    EgoWM (Cosmos) &  &  & 480$\times$640 & $\approx 50$ \\
    \hline
    \end{tabular}}
\end{table*}
\begin{figure*}[t]
    \centering
    \includegraphics[width=\linewidth]{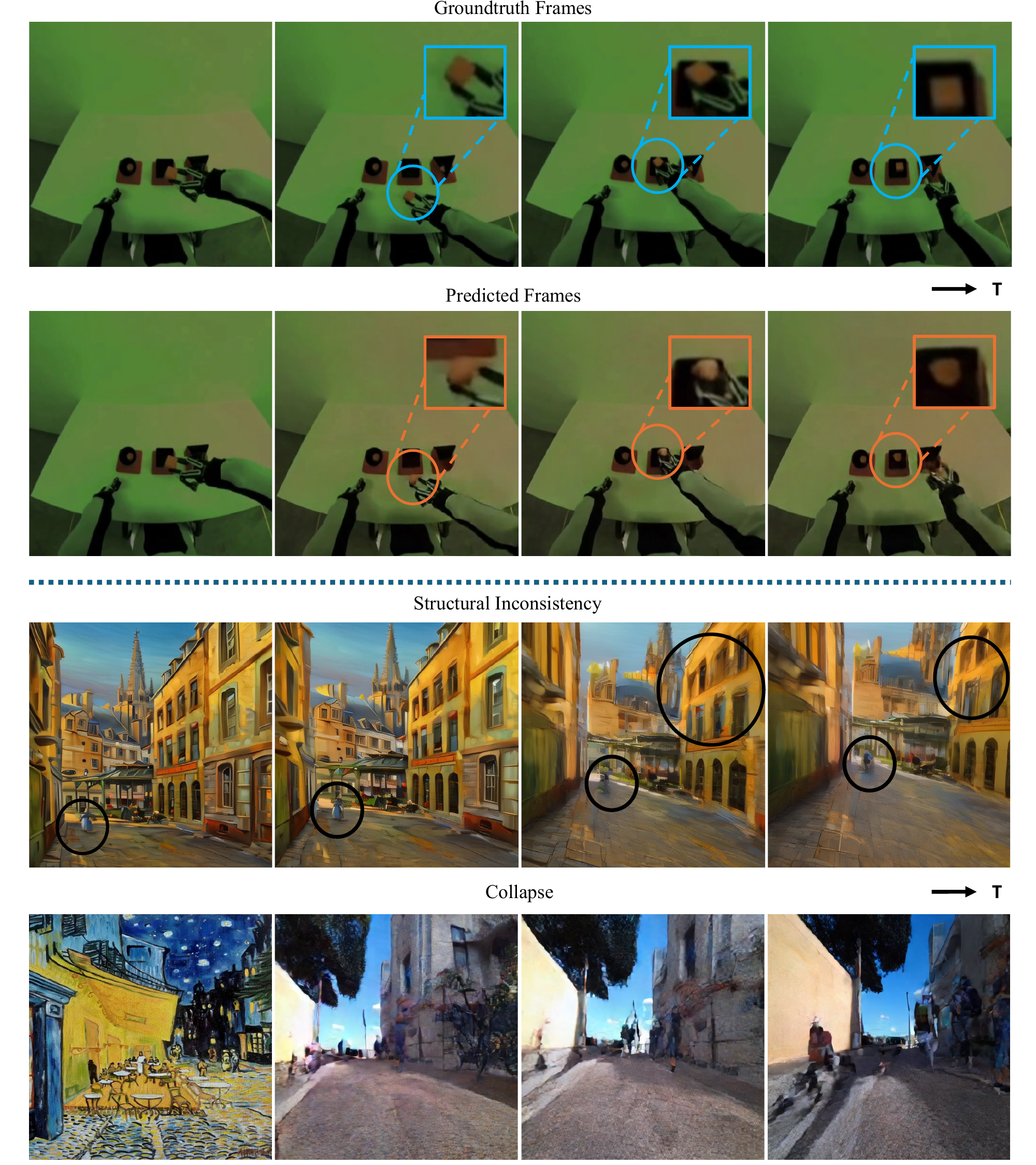}
    \caption{Failure modes of our world model. In the first two rows we show shape inconsistencies in our manipulation results, in the third row we show structural inconsistencies in our SVD 3-DoF navgiation world model's painting generalizations and in the last row we show collapse to a real-world scene in our Cosmos 3-DoF navigation world model.}
    \label{fig:failure}
\end{figure*}

\section{Implementation Details}
\label{sec:impl_supp}
We train both our SVD~\cite{blattmann2023stable} and Cosmos~\cite{agarwal2025cosmos} models on 8 A100 GPUs for all variants. For the SVD variant, we set the learning rate to 1e-5 for all the model parameters except the action projection layers, where we adopt a larger learning rate of 1e-4. For the Cosmos and Wan variants, we use a learning rate of 1e-6 for all the parameters except for the action projection layers, which use learning rate of 1e-5. For the Wan model we keep all weights at bfloat16 precision except for the newly introduced action module's weights and the base model's timestep conditioning  weights (including the scale/shift/gate parameters), which are set to float32. 

We subsample the frames at 5 FPS for both training and inference. For 1X training, we select continuous frame sequences with a stride of 5. For training on 3-DoF datasets, we use a stride of 1 for SCAND~\cite{karnan2022socially} and Tartan~\cite{triest2022tartandrive}, and a stride of 5 for RECON. The SVD model is trained at 512×512 resolution to predict 8 future frames and evaluated autoregressively over two 8-frame chunks (16 frames total), while the Cosmos and Wan variants operate at 480×640 resolution and directly predict 16 frames in a single forward pass.

\section{Failure Modes}
\label{sec:failure}

\noindent\textbf{Manipulation.} Our model sometimes struggles to generate small manipulated objects in a physically consistent manner, particularly when object shapes must be preserved across occlusions. Maintaining object permanence during complex manipulation also remains challenging. In the first two rows of Fig~\ref{fig:failure}, we illustrate one such failure case, where the square shaped block gets distorted during manipulation. We attribute these issues primarily to limitations of the underlying video generation backbones and expect that future advances in video generative modeling will improve these aspects of world-modeling performance.

\noindent\textbf{Generalization.}
While our model demonstrates strong generalization to highly out-of-distribution paintings, we observe two dominant failure modes. First, in some cases the generated scene collapses toward the real-world training distribution, reducing stylistic consistency with the input painting (see last row of Fig.~\ref{fig:failure}). Second, structural inconsistencies can emerge during generation. When navigation commands require rapid synthesis of previously unseen regions, the model may lose scene fidelity or fail to maintain consistent representations of people and structures over time. This can manifest as objects or architectural elements vanishing, appearing, or morphing across frames. An example of such structural instability is shown in the third row of Fig.~\ref{fig:failure}. We expect these limitations to diminish as base video models improve in long-horizon coherence and structural consistency. However, improving structural consistency using post-training techniques remains an important direction for future work.

\end{document}